\documentclass[10pt,twocolumn,letterpaper]{article}

\usepackage[pagenumbers]{cvpr} 

\usepackage{multirow}
\usepackage{mathtools}

\usepackage{microtype}

\renewcommand{\paragraph}[1]{\vspace{.5em}\noindent\textbf{#1.}}

\definecolor{cvprblue}{rgb}{0.21,0.49,0.74}
\usepackage[pagebackref,breaklinks,colorlinks,allcolors=cvprblue]{hyperref}

\def\confName{CVPR}
\def\confYear{2027}

\title{FoRe: A Force-Guided Framework for Autonomous Grasp Recovery in Vision-Language-Action Models}

\author{Shuai Peng \qquad Yongkai Liao \qquad Yuxin Cheng \qquad Haoyuan Wang \qquad Jian Huang}

\begin{document}
\maketitle
\begin{abstract}
Vision-language-action (VLA) models have demonstrated strong performance in grasp-based manipulation, yet reliable recovery from local grasping failures remains challenging. Specifically, in pick-and-place tasks, policies may continue transport after a failed grasp instead of retrying, potentially reflecting overreliance on proprioceptive cues for stage progression and limited retry supervision. Existing recovery approaches typically rely on deployment-time guidance or demonstration augmentation. We introduce Force-guided Recovery (FoRe), a framework for autonomous recovery from local grasping failures. Using force feedback, FoRe partitions demonstrations into approach–grasp and transport–placement stages to train separate policies. To supplement limited retry supervision, FoRe then initializes a dedicated retry policy from the approach–grasp policy and refines it through local reinforcement learning (RL) from post-failure states. To avoid inappropriate stage progression, a controller combines force feedback with grasp-command status at inference to coordinate the stage policies, detect grasp failures, and invoke the retry policy. We also construct LIBERO-SUCTION, a force-augmented suction manipulation benchmark adapted from LIBERO, with complete demonstration trajectories for policy training. FoRe consistently improves task success rates for VLA-Adapter and $\pi_{0.5}$ on LIBERO-SUCTION, LIBERO, and real-world tasks. On LIBERO-SUCTION, average absolute gains are 18.4\% and 6.5\%, respectively. The corresponding real-world absolute gains are 25.0\% and 21.7\%. FoRe also remains effective with only one demonstration per task on both simulation benchmarks.
\end{abstract}    
\section{Introduction}

Vision-language-action (VLA) models leverage pretrained visual-language knowledge and diverse robot data for generalizable manipulation~\cite{RT2,OpenVLA,OXE,BridgeV2,DROID}, with advances in action generation, spatial understanding, reasoning, and long-horizon manipulation~\cite{pi0,pi05,OpenVLA-OFT,SpatialVLA,CoTVLA}. However, especially in pick-and-place tasks, reliable recovery from local execution errors remains challenging: an imitation-trained VLA policy may continue transport after a failed grasp, instead of retrying~\cite{RLGeneralization,FLARE}. Such behavior can turn a recoverable local error into overall task failure, motivating autonomous retries and task resumption.

Two related limitations may explain why VLA policies fail to recover after an unsuccessful grasp. First, as shown in Fig.~\ref{fig:question}, manipulation tasks can be viewed as sequences of stages, such as the approach–grasp stage and the transport–placement stage, each with its own distinct objective.  Because robot proprioceptive states often correlate with task progress in successful demonstrations, a policy may rely excessively on these proprioceptive state cues to transition to the corresponding stage, regardless of the actual grasp outcome~\cite{CausalConfusion,FLARE}. Consequently, after a failed grasp, a robot proprioceptive state resembling those associated with transport may prompt the policy to persist in generating transport actions rather than retrying the grasp. Second, training dominated by uninterrupted successful demonstrations provides limited supervision for interrupting nominal execution and re-approaching the object after failure~\cite{FLARE}. The policy may therefore lack an effective response even when failure cues are observable.

\begin{figure*}[t]
  \centering
  \includegraphics[width=\textwidth]{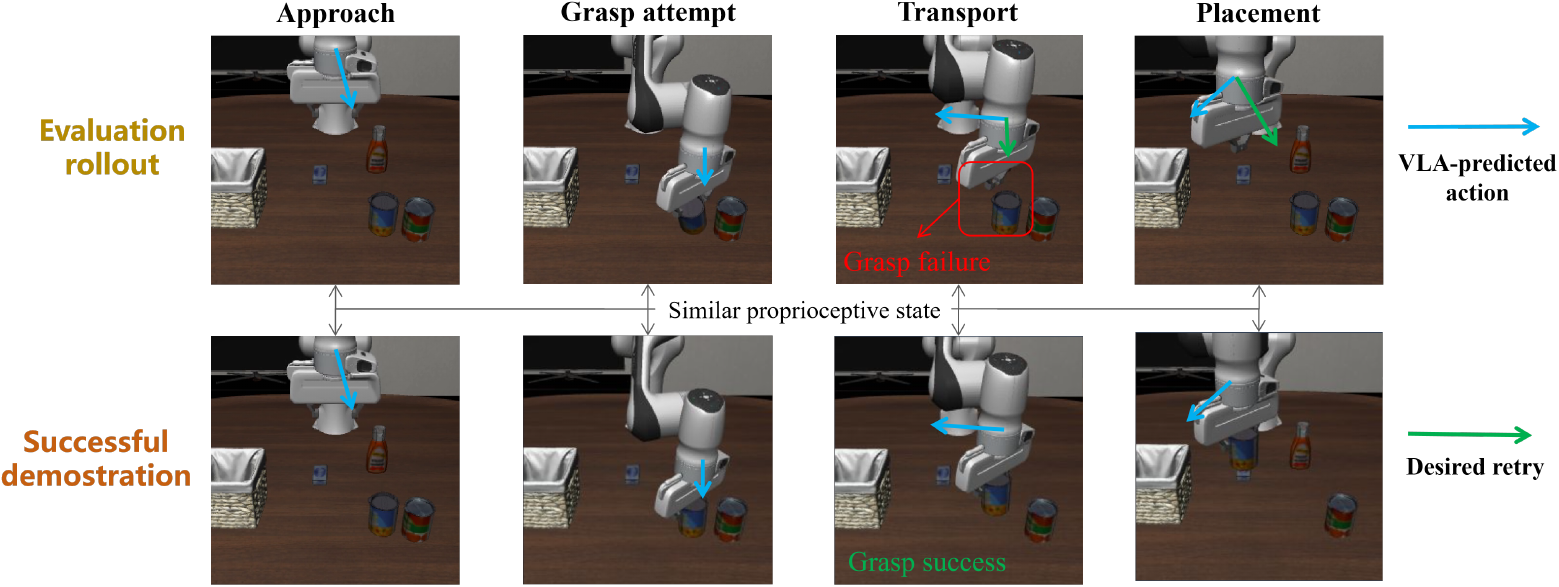}
  \caption{Continued transport after grasp failure. During evaluation (top), the VLA policy continues transport while the object remains on the table. Its post-failure state resembles transport-stage states in the successful demonstration (bottom), prompting continued transport despite failure through overreliance on proprioceptive cues. Blue arrows indicate VLA-predicted actions; green arrows indicate desired retries.}
  \label{fig:question}
\end{figure*}

Recent studies have explored several approaches for robotic recovery. One direction uses human feedback~\cite{RTH,YellRobot,OLAF} or vision-language model (VLM) guidance~\cite{AHA,FailSafe,ViFailback,CycleVLA} to monitor execution outcomes and guide stage progression. Another direction augments demonstrations~\cite{FLARE} with varied robot poses and corrective trajectories to weaken correlations between pose and task progress and expand retry supervision. Force feedback provides direct physical cues about contact and load changes, making it an effective signal for detecting grasp failures. However, an explicit mechanism that uses such feedback to coordinate grasp retries and task resumption in VLA policies remains underexplored.

We introduce Force-guided Recovery (FoRe), a framework for autonomous recovery from local grasping failures in VLA-based manipulation. FoRe first uses force feedback to partition successful demonstrations into approach–grasp and transport–placement stages, training a separate policy for each. To supplement limited retry supervision, FoRe then initializes a dedicated retry policy from the approach–grasp policy and refines it through local reinforcement learning (RL) from post-failure states. To avoid inappropriate stage progression, a controller combines force feedback with grasp-command status at inference to coordinate the stage policies, detect grasp failures, and invoke the retry policy. FoRe thus mitigates the adverse effects of relying on proprioceptive cues for stage progression through explicit stage control, while local retry learning supplements limited recovery supervision. Our main contributions are threefold:

\begin{itemize}
  \item We propose FoRe, a framework that integrates independent training of two stage-specific policies, local refinement of a dedicated retry policy, and force-guided stage control to enable autonomous grasp recovery across VLA backbones.
  \item To support research on autonomous grasp recovery in suction-based manipulation, we construct LIBERO-SUCTION, a force-augmented benchmark adapted from LIBERO, comprising 22 tasks across three suites and 1,100 complete demonstrations for VLA training. The benchmark and demonstrations will be publicly released.
  \item We demonstrate that FoRe yields consistent improvements in task success with VLA-Adapter and $\pi_{0.5}$, on LIBERO-SUCTION, LIBERO, and real-world tasks. Further evaluations show that FoRe also remains effective with only one demonstration per task on both simulation benchmarks.
\end{itemize}

\section{Related Work}

\subsection{Vision-Language-Action Models}

Vision-language-action (VLA) models have substantially advanced robotic manipulation by transferring pretrained visual and linguistic knowledge to action generation~\cite{RT2}. OpenVLA~\cite{OpenVLA} demonstrates strong language grounding and adaptation to new tasks, while OpenVLA-OFT~\cite{OpenVLA-OFT} further improves task performance and inference efficiency. $\pi_0$~\cite{pi0} and $\pi_{0.5}$~\cite{pi05} extend VLA capabilities to dexterous control and long-horizon manipulation in diverse environments. SpatialVLA~\cite{SpatialVLA} incorporates 3D spatial representations to improve spatial understanding, while CoT-VLA~\cite{CoTVLA} introduces visual chain-of-thought reasoning by predicting intermediate goal images before action generation.

Despite their strong generalization within the training task distribution, these models have limited autonomous recovery capabilities. Consequently, minor execution errors can escalate into task failures, substantially compromising the reliability of VLA task execution.

\begin{figure*}[t]
  \centering
  \includegraphics[width=\textwidth]{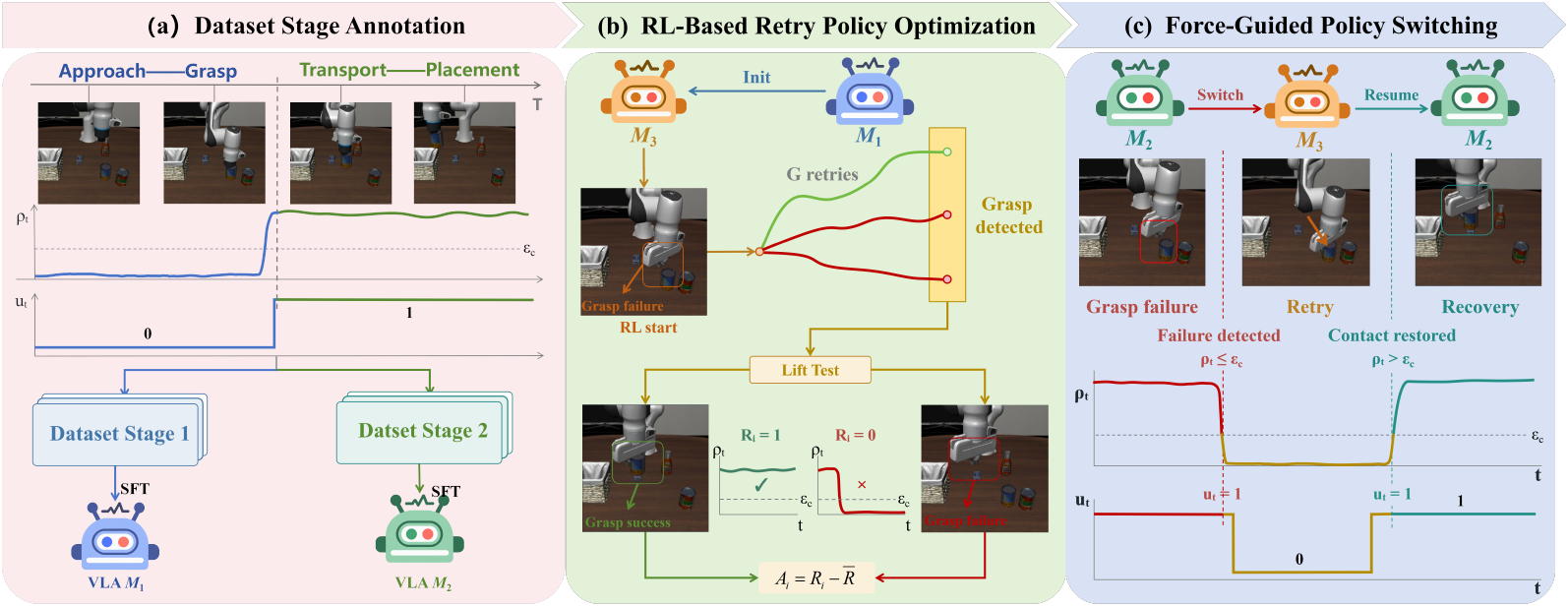}
  \caption{Overview of FoRe. (a) Force feedback and grasp commands partition successful demonstrations into approach--grasp and transport--placement stages for independently fine-tuning \(M_1\) and \(M_2\). (b) Initialized from \(M_1\), \(M_3\) is refined by local RL from simulated post-failure states. For \(G\) retries with the same initial state and instruction, lift tests yield binary rewards and group-relative advantages \(A_i=R_i-\bar{R}\). (c) Contact loss under an active grasp command triggers \(M_2\!\rightarrow\!M_3\); restored contact resumes \(M_2\) for transport and placement.}
  \label{fig:method}
\end{figure*}

\subsection{Failure Recovery in Robotic Manipulation}

Human-guided approaches use language feedback to redirect execution and improve robot policies. RT-H~\cite{RTH} supports interventions through language descriptions of low-level motions, while Yell at Your Robot~\cite{YellRobot} incorporates real-time verbal corrections into hierarchical policies. OLAF~\cite{OLAF} uses verbal corrections to relabel actions for policy updates. Obtaining such corrective feedback requires human participation, which can limit scalability. VLM-based approaches automate failure analysis and recovery guidance: FailSafe~\cite{FailSafe} predicts  recovery actions, and CycleVLA~\cite{CycleVLA} combines VLM-guided subtask backtracking with minimum Bayes risk decoding for retries. FLARE~\cite{FLARE} combines learned retries with reset skills invoked by an online multimodal monitor, while ReCoVLA~\cite{ReCoVLA} uses VLM-based failure detection to activate residual recovery policies. Online VLM supervision can introduce additional inference overhead and latency~\cite{CycleVLA}.

In contrast to these approaches, FoRe introduces a novel and physically grounded approach to autonomous grasp recovery, using force feedback and grasp-command status to govern stage transitions, detect grasp failures and activate a retry policy refined through local RL.

\subsection{Reinforcement Learning for VLA Models}

RL improves VLA policies through interaction and task-outcome feedback, with gains in performance and execution generalization~\cite{SimpleVLA,RLGeneralization}. ConRFT and RECAP study RL-based post-training for real-world policy improvement~\cite{ConRFT,RECAP}. PLD trains residual specialists to collect recovery trajectories for fine-tuning the base policy~\cite{PLD}. SRPO uses model-generated successful trajectories as references for progress-based rewards on failed attempts~\cite{SRPO}.

Compared to the above works, FoRe incorporates RL as a local refinement step within a broader recovery framework, optimizing only a dedicated retry policy from post-failure states.

\section{Methodology}

\subsection{Problem Formulation}
We consider language-conditioned robotic manipulation. Given visual observation \(v_t\), robot proprioceptive state \(x_t\), and language instruction \(\ell\), a nominal VLA policy generates actions as \(a_t\sim\pi_\theta(\cdot\mid v_t,x_t,\ell)\). The policy is trained by imitation learning on successful demonstrations \(\mathcal D_{\mathrm{demo}}=\{\tau_i\}_{i=1}^{N}\).

Let \(\mathcal X_{\mathrm{tr}}^{\mathrm{demo}}\) denote robot proprioceptive states recorded during transport in successful demonstrations of the same task. After a failed grasp, the current proprioceptive state may still resemble one in this set. A policy that overrelies on proprioceptive state cues may consequently favor the associated transport actions over retry actions. One possible manifestation of this failure is
\begin{equation}
\begin{gathered}
x_t \approx x_j^{\mathrm{demo}},
\qquad x_j^{\mathrm{demo}}\in\mathcal X_{\mathrm{tr}}^{\mathrm{demo}},\\
\pi_\theta(\mathcal A_{\mathrm{tr}}\mid v_t,x_t,\ell)
>
\pi_\theta(\mathcal A_{\mathrm{retry}}\mid v_t,x_t,\ell),
\end{gathered}
\end{equation}
where \(\mathcal A_{\mathrm{tr}}\) and \(\mathcal A_{\mathrm{retry}}\) denote transport and retry actions, respectively. Moreover, training on successful demonstrations provides limited supervision for retry after failed grasps.

\subsection{Framework Overview}
The FoRe aims to interrupt inappropriate task progression after a failed grasp and enable autonomous grasp retries through physical interaction feedback. Using force-guided stage annotation (Sec. 3.3), we partition \(\mathcal D_{\mathrm{demo}}\) into approach–grasp and transport–placement subsets, then independently fine-tune the approach–grasp policy \(M_1\) and the transport–placement policy \(M_2\) on their respective subsets, as illustrated in Fig.~\ref{fig:method}(a).

As shown in Fig.~\ref{fig:method}(b), we initialize the retry policy \(M_3\) from \(M_1\) and refine it through local RL focused on grasp retries (Sec. 3.4).

At inference, a force-guided controller (Sec. 3.5) uses force measurements to select \(M_1\) for nominal approach and grasp, \(M_2\) for transport and placement, or \(M_3\) for retries following detected grasp failures, as shown in Fig.~\ref{fig:method}(c).

\subsection{Dataset Stage Annotation}
Initial grasp attempts and retries share the local objective of approaching and grasping the target. To reuse this behavior for retry learning, we assign each demonstration time step a stage label \(c_t\in\{1,2\}\), corresponding to approach–grasp (Stage 1) and transport–placement (Stage 2), respectively. 

Robot motion introduces inertial contributions into force measurements. We use robot acceleration signals to reduce these contributions and denote the force by $\mathbf f_t$, with magnitude $\rho_t=\|\mathbf f_t\|_2$. We then determine the detection threshold $\epsilon_c$ separately for each dataset through manual statistical analysis of its force signals. The threshold is selected to distinguish residual no-load fluctuations from contact and load signals. Within each setup, the threshold remains fixed and is shared by demonstration annotation, retry termination, and runtime policy switching. Let $u_t\in\{0,1\}$ denote the grasp command, with $0$ indicating release and $1$ indicating grasp activation. For $t\ge1$, we define
\begin{equation}
\begin{aligned}
g_t&=\mathbf1\!\left[
\rho_t>\epsilon_c
\;\land\;u_t=1
\right],\\
r_t&=\mathbf1\!\left[
\rho_t\le\epsilon_c
\;\land\;u_t=0
\right].
\end{aligned}
\end{equation}
Initializing \(c_0=1\), we update the stage label as
\begin{equation}
c_t=
\begin{cases}
2, & c_{t-1}=1,\ g_t=1,\\
1, & c_{t-1}=2,\ r_t=1,\\
c_{t-1}, & \text{otherwise}.
\end{cases}
\end{equation}
The resulting labels partition \(\mathcal D_{\mathrm{demo}}\) into Stage 1 (\(c_t=1\)) and Stage 2 (\(c_t=2\)) subsets for training VLA \(M_1\) and VLA \(M_2\), respectively, as shown in Fig.~\ref{fig:method}(a).

\subsection{RL-Based Retry Policy Optimization}

Successful demonstrations provide limited supervision for retries after failed grasps. To learn the retry behavior, we initialize the retry policy $M_3$ from $M_1$ to reuse the approach–grasp behavior learned through stage-specific training. We then refine $M_3$ through local RL in grasp-retry scenarios to adapt this behavior to states encountered after failed grasps.

To construct grasp-retry scenarios, we sample robot poses from an early temporal window of each Stage 2 demonstration segment. These early transport poses resemble those the robot may assume after a failed grasp. During initialization, we set the robot to the sampled pose and leave the target object at its original, ungrasped pose. The resulting scene simulates the state after a failed grasp and serves as the starting point for optimizing \(M_3\) through RL.

For each initialized scenario, as shown in Fig.~\ref{fig:method}(b), we collect $G$ stochastic rollouts sharing the same initial state and instruction $\ell$. Let $o_{i,t}=(v_{i,t},x_{i,t},\ell)$ denote the policy input. To explore alternative retry behaviors, we tailor stochastic sampling to the action-generation mechanism of each backbone. For flow-based action generation, represented by $\pi_{0.5}$~\cite{pi05}, we sample Gaussian initial noise and iteratively integrate the learned flow to generate action chunks. For deterministic action regression, represented by VLA-Adapter~\cite{vla-adapte}, we introduce Gaussian exploration around the action predicted by $M_3$:
\begin{equation}
\widetilde{\pi}_\theta(\mathbf a\mid o)
=
\mathcal{N}\bigl(
\mathbf a;\boldsymbol{\mu}_\theta(o),\boldsymbol{\Sigma}
\bigr),
\end{equation}
where $\boldsymbol{\mu}_\theta(o)$ is the predicted action and $\boldsymbol{\Sigma}$ is a fixed exploration covariance.

We assign a binary RL reward based on a lift test. Specifically, we terminate the retry under the joint condition $\rho_t>\epsilon_c \land u_{i,t}=\mathrm{ON}$, which indicates an active grasp attempt. The subsequent lift test verifies grasp success by checking whether the force signal remains stable above the near-zero no-load baseline during lifting. A rapid drop followed by fluctuations near zero indicates failure. We assign $R_i=1$ if the grasp attempt passes the lift test, and $R_i=0$ otherwise.

To compare alternative retry from the same initial state, we compute the group mean reward and a trajectory-level relative advantage:
\begin{equation}
\bar R=\frac{1}{G}\sum_{i=1}^{G}R_i,
\qquad
A_i=R_i-\bar R.
\end{equation}
All retry actions within trajectory $i$ share the same $A_i$. In mixed-outcome groups, successful attempts receive positive advantages and failed attempts receive negative advantages. Groups with identical outcomes have zero relative advantages and do not contribute to the update.

Treating the collected advantages as constants, we update $M_3$ by minimizing the trajectory-weighted surrogate loss:
\begin{equation}
\mathcal L_{\mathrm{RL}}(\theta)
=
-\frac{1}{Z}
\sum_{i=1}^{G}A_i
\sum_{t=0}^{T_i-1}
\log\widetilde\pi_\theta
\bigl(\mathbf a_{i,t}\mid o_{i,t}\bigr).
\end{equation}
Here, $Z=\sum_{i=1}^{G}T_i$ is the total number of retry actions in the group.

\subsection{Force-Guided Policy Switching}
At inference, as shown in Fig.~\ref{fig:method}(c), the controller combines force measurements with grasp-command status to switch among three policies: $M_1$ for nominal approach and grasp, $M_2$ for transport and placement, and $M_3$ for retries following detected grasp failures.

\paragraph{Nominal stage transitions}
During nominal execution, the controller starts with $M_1$ for routine approach and grasp, and switches to $M_2$ for transport and placement once $u_t=1$ and the force magnitude $\rho_t$ exceeds $\epsilon_c$, indicating contact during an active grasp attempt. After placement, a release command ($u_t=0$) indicates completion of the release, upon which the controller returns to $M_1$ for the next grasp cycle if the task continues.

\paragraph{Grasp-failure detection}
If the grasp succeeds, the force magnitude remains approximately stable at a nonzero level during transport under $M_2$. By contrast, grasp failure causes the signal to drop rapidly toward zero and fluctuate near the no-load baseline, although the grasp command remains active. We therefore detect grasp failure during $M_2$ execution by combining the $\epsilon_c$ with the grasp-command status:
\begin{equation}
f_t=
\begin{cases}
\mathbf{1}\bigl[u_t=1\land\rho_t\le\epsilon_c\bigr],
& \text{if }M_2\text{ is active},\\
0,
& \text{otherwise}.
\end{cases}
\end{equation}
The active grasp command distinguishes this loss of contact from intentional release after placement.

\paragraph{Unified policy switching}
The controller starts execution with $M_1$ and switches among the policies according to the following rules:
\begin{equation}
M_1
\xrightleftharpoons[
\,u_t=0
]{
\,u_t=1,\ \rho_t>\epsilon_c\,
}
M_2
\xrightleftharpoons[
\,u_t=1,\ \rho_t>\epsilon_c\,
]{
\,f_t=1\,
}
M_3.
\end{equation}
At each timestep, only transitions from the active policy are evaluated, and the current policy is retained unless a transition condition is satisfied. This switching strategy supports accurate and rapid nominal execution by coordinating $M_1$ and $M_2$, while enabling real-time detection of grasp failures and efficient recovery through retries with $M_3$.

\section{Experiments}

We evaluate whether the FoRe improves end-to-end task completion across manipulation tasks and VLA backbones. Our experiments address two questions:

\begin{itemize}[label={},leftmargin=1.5em,itemindent=-1em,labelsep=0pt,itemsep=0.2\baselineskip,topsep=0.2\baselineskip,parsep=0pt]
  \item \textit{\textbf{Question 1.} How does the FoRe affect task success relative to VLA baselines in simulation and on a physical robot? (Secs. 4.1 and 4.2)}
  \item \textit{\textbf{Question 2.} Does FoRe remain effective with one trajectory per task? (Sec. 4.3)}
\end{itemize}

We evaluate FoRe on LIBERO-SUCTION, LIBERO~\cite{LIBERO}, and real-world tasks, including one-demonstration training on both simulation benchmarks. Experiments use a server with eight NVIDIA 4090 GPUs.

\begin{table*}[!t]
  \centering
  \caption{Task success rates (\%) on LIBERO-SUCTION (higher is better). Bold and underlining mark the best and second-best results. “w/ FoRe” applies the FoRe to the backbone immediately above it; red $\Delta$ values are absolute gains in percentage points.}
  \label{tab:libero_suction_results}
  \setlength{\tabcolsep}{8pt}
  \renewcommand{\arraystretch}{0.9}
  \begin{tabular*}{\textwidth}{@{\extracolsep{\fill}}lcccc}
    \toprule
    \multirow{2}{*}{\textbf{Model}} &
    \multicolumn{4}{c}{\textbf{LIBERO-SUCTION}} \\
    \cmidrule(lr){2-5}
    &
    \shortstack{\textbf{Suction-Spatial}} &
    \shortstack{\textbf{Suction-Object}} &
    \shortstack{\textbf{Suction-Goal}} &
    \textbf{Avg} \\
    \midrule
    OpenVLA        & 69.3 & 50.5 & 66.8 & 61.9\\
    OpenVLA-OFT    & 82.2 & 66.0 & 83.2 & 76.5\\
    $\pi_{0}$      & 79.1 & 61.3 & 82.0 & 73.3\\
    $\pi_{0}$-FAST & 84.2 & 66.3 & 81.2 & 77.0\\
    SmolVLA        & 80.4 & 63.0 & 79.6 & 73.9\\
    GR00T N1       & 81.3 & 65.0 & 82.4 & 75.6\\
    \midrule
    VLA-Adapter    & 77.3 & 52.3 & 81.6 & 69.2\\
    \quad\textbf{w/ FoRe} & 91.3 & \underline{78.8} & 95.2 & \underline{87.6}\\
    \quad$\Delta$  & \textcolor{red!70!black}{+14.0} & \textcolor{red!70!black}{+26.5} & \textcolor{red!70!black}{+13.6} & \textcolor{red!70!black}{+18.4}\\
    \addlinespace[3pt]
    $\pi_{0.5}$    & \underline{94.0} & 70.3 & \underline{96.4} & 85.9\\
    \quad\textbf{w/ FoRe} & \textbf{98.9} & \textbf{81.8} & \textbf{97.6} & \textbf{92.4}\\
    \quad$\Delta$  & \textcolor{red!70!black}{+4.9} & \textcolor{red!70!black}{+11.5} & \textcolor{red!70!black}{+1.2} & \textcolor{red!70!black}{+6.5}\\
    \bottomrule
  \end{tabular*}
\end{table*}

\begin{table*}[t]
  \centering
  \caption{Task success rates (\%) on LIBERO (higher is better). Bold and underlining mark the best and second-best results. “w/” denotes applying the method to the corresponding backbone. Parentheses indicate absolute gains over the corresponding base policy.}
  \label{tab:libero_results}
  \setlength{\tabcolsep}{8pt}
  \renewcommand{\arraystretch}{0.9}
  \begin{tabular*}{\textwidth}{@{\extracolsep{\fill}}lccccc}
    \toprule
    \multirow{2}{*}{\textbf{Model}} &
    \multicolumn{5}{c}{\textbf{LIBERO}} \\
    \cmidrule(lr){2-6}
    &
    \shortstack{\textbf{Spatial}} &
    \shortstack{\textbf{Object}} &
    \shortstack{\textbf{Goal}} &
    \shortstack{\textbf{Long}} &
    \textbf{Avg} \\
    \midrule
    OpenVLA            & 84.7 & 88.4 & 79.2 & 53.7 & 76.5\\
    OpenVLA-OFT        & 97.6 & 98.4 & 97.9 & 94.5 & 97.1\\
    $\pi_{0}$-FAST     & 96.4 & 96.8 & 88.6 & 60.2 & 85.5\\
    GR00T N1           & 94.4 & 97.6 & 93.0 & 90.6 & 93.9\\
    \midrule
    VLA-Adapter        & 92.4 & 92.6 & 91.2 & 87.8 & 91.0\\
    \quad{w/ CycleVLA} & 98.2 & 99.0 & 97.8 & 95.4 & 97.6 (+6.6)\\
    \quad\textbf{w/ FoRe} & \underline{99.2} & \textbf{99.6} & \textbf{98.2} & \textbf{97.4} & \textbf{98.6} (+7.6)\\
    \addlinespace[3pt]
    $\pi_{0.5}$         & 93.8 & 92.2 & 92.8 & 89.4 & 92.1\\
    \quad{w/ CycleVLA}  & 98.8 & 99.4 & 97.6 & 95.8 & 97.9 (+5.8)\\
    \quad\textbf{w/ FoRe} & \textbf{99.6} & \underline{99.4} & \underline{98.0} & \underline{96.2} & \underline{98.3} (+6.2)\\
    \bottomrule
  \end{tabular*}
\end{table*}

\subsection{Performance on Simulation Benchmarks}
\paragraph{LIBERO-SUCTION Benchmark}
We construct LIBERO-SUCTION to study autonomous grasp recovery in suction manipulation, adapting LIBERO with a suction gripper and force feedback for monitoring physical interaction. Its 22 tasks span Suction-Spatial, Suction-Object, and Suction-Goal, assessing spatial relations, object identification, and goal-directed execution, respectively. The benchmark includes 1,100 complete demonstration trajectories for VLA training, matched to the evaluation environments.

\paragraph{LIBERO Benchmark}
We also evaluate FoRe on four LIBERO task suites—Spatial, Object, Goal, and Long—using the original gripper and environments augmented with force feedback. Together, the two benchmarks cover parallel-jaw and suction-based grasping.

\paragraph{Baselines} We compare eight VLA baselines: OpenVLA~\cite{OpenVLA}, OpenVLA-OFT~\cite{OpenVLA-OFT}, $\pi_{0}$~\cite{pi0}, $\pi_{0}$-FAST~\cite{pi0-fast}, SmolVLA~\cite{smolvla}, GR00T N1~\cite{gr00tn1}, VLA-Adapter~\cite{vla-adapte}, and $\pi_{0.5}$~\cite{pi05}. We evaluate FoRe on VLA-Adapter and $\pi_{0.5}$, comparing its performance with CycleVLA~\cite{CycleVLA}.

\paragraph{Evaluation protocol} Each model is evaluated over 50 episodes per task: 1,100 episodes across 22 tasks in three LIBERO-SUCTION suites, and 2,000 episodes across 40 tasks in four LIBERO suites. We report success rates (\%) per suite and across all evaluated tasks in each benchmark.

\begin{figure}[!htbp]
  \centering
  \includegraphics[width=\linewidth]{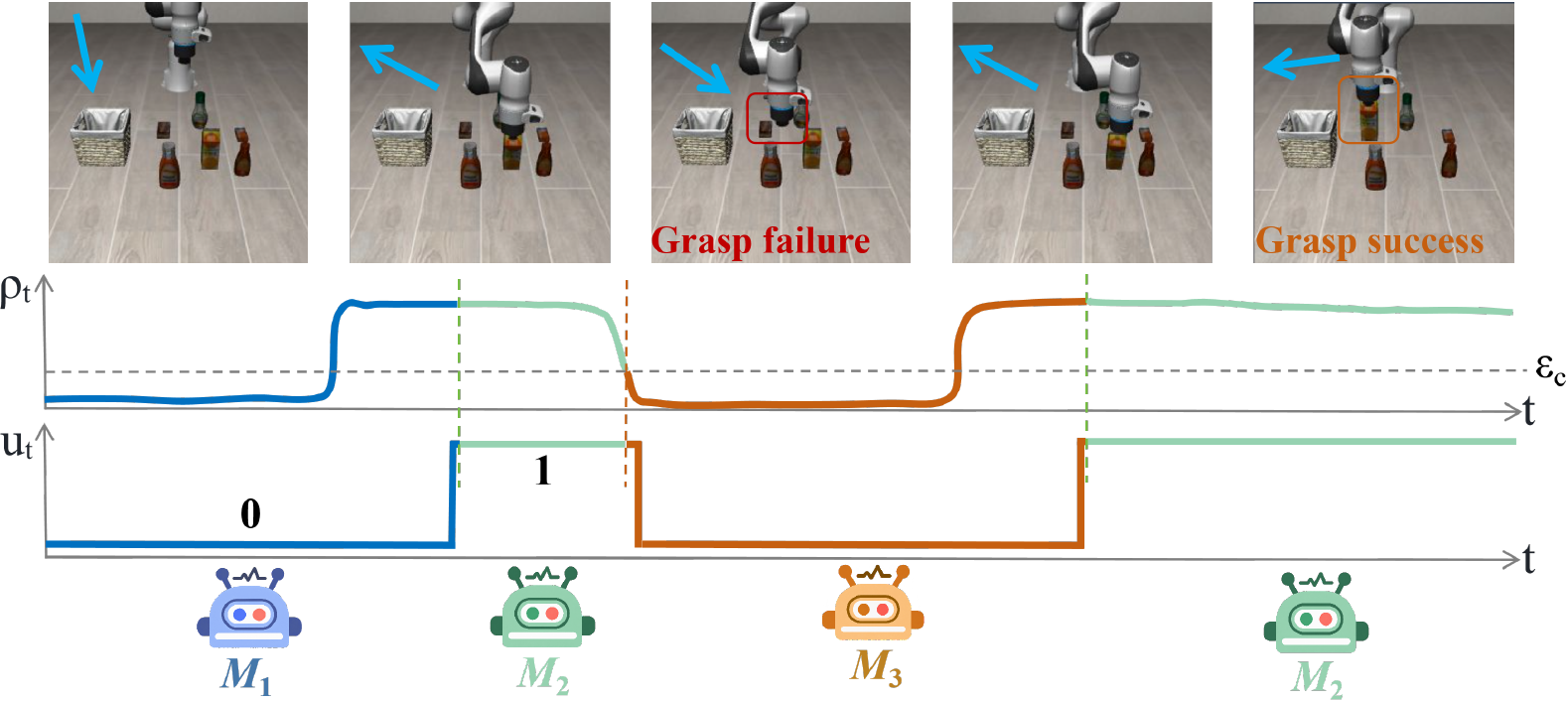}
  \caption{Inference-time grasp recovery on LIBERO-SUCTION. Force feedback $\rho_t$ and grasp-command status $u_t$ guide policy switching ($M_1 \rightarrow M_2 \rightarrow M_3 \rightarrow M_2$), enabling autonomous recovery from grasping failures. Blue arrows indicate VLA-predicted actions.}
  \label{fig:LIBERO-SUCTION-visual}
\end{figure}

\paragraph{Results on LIBERO-SUCTION} Table~\ref{tab:libero_suction_results} shows that the FoRe improves the overall success rate of VLA-Adapter from 69.2\% to 87.6\%, an absolute gain of 18.4\%. For $\pi_{0.5}$, the success rate increases from 85.9\% to 92.4\%, an absolute gain of 6.5\%. This is the highest overall score among the evaluated models in this table. The recovery sequence is illustrated in Fig.~\ref{fig:LIBERO-SUCTION-visual}. These results show that the FoRe improves task completion across the evaluated task groups and with both tested action-generation mechanisms. 

\begin{figure}[!htbp]
  \centering
  \includegraphics[width=\linewidth]{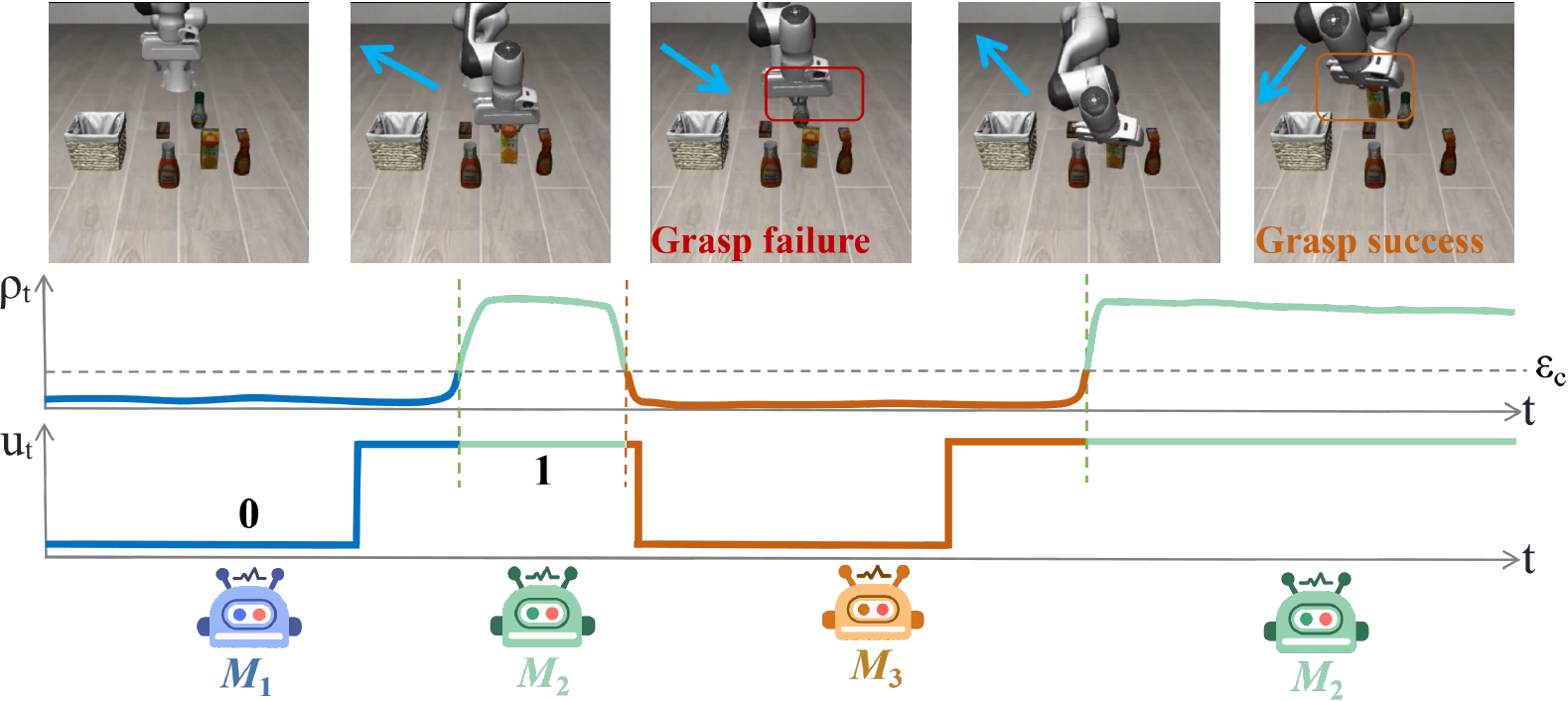}
  \caption{Inference-time grasp recovery on LIBERO. All details are as in Fig.~\ref{fig:LIBERO-SUCTION-visual}.}
  \label{fig:LIBERO-visual}
\end{figure}

\paragraph{Results on LIBERO} Table~\ref{tab:libero_results} compares FoRe with the base policies and CycleVLA across four LIBERO suites. FoRe achieves average success rates of 98.6\% and 98.3\% for VLA-Adapter and $\pi_{0.5}$, respectively, corresponding to absolute gains of 7.6\% and 6.2\% over the base policies. FoRe also outperforms CycleVLA for both backbones, with average absolute gains of 1.0\% and 0.4\%, respectively. Fig.~\ref{fig:LIBERO-visual} illustrates FoRe's policy switching and grasp recovery during inference on LIBERO.

\begin{table*}[t]
  \centering
  \caption{Task success rates (\%) on LIBERO-SUCTION with one demonstration per task (higher is better). “w/ FoRe” applies the FoRe under the corresponding demonstration regime; $\Delta$ is the absolute gain over the SFT baseline in percentage points.
}
  \label{tab:libero_suction_one_trajectory}
  \setlength{\tabcolsep}{8pt}
  \renewcommand{\arraystretch}{0.9}
  \begin{tabular*}{\textwidth}{@{\extracolsep{\fill}}lcccc}
    \toprule
    \multirow{2}{*}{\textbf{Model}} &
    \multicolumn{4}{c}{\textbf{LIBERO-SUCTION}} \\
    \cmidrule(lr){2-5}
    &
    \shortstack{\textbf{Suction-Spatial}} &
    \shortstack{\textbf{Suction-Object}} &
    \shortstack{\textbf{Suction-Goal}} &
    \textbf{Avg} \\
    \midrule
    \multicolumn{5}{c}{One-Trajectory SFT}\\
    \midrule
    VLA-Adapter           & 42.0 & 23.8 & 30.8 & 32.8\\
    \quad\textbf{w/ FoRe} & 66.9 & 43.8 & 53.6 & 55.5\\
    \quad$\Delta$  & \textcolor{red!70!black}{+24.9} & \textcolor{red!70!black}{+20.0} & \textcolor{red!70!black}{+22.8} & \textcolor{red!70!black}{+22.7}\\
    \addlinespace[3pt]
    $\pi_{0.5}$           & 48.4 & 30.5 & 36.4 & 39.2\\
    \quad\textbf{w/ FoRe} & \textbf{72.0} & \textbf{56.3} & \textbf{58.8} & \textbf{63.3}\\
    \quad$\Delta$  & \textcolor{red!70!black}{+23.6} & \textcolor{red!70!black}{+25.8} & \textcolor{red!70!black}{+22.4} & \textcolor{red!70!black}{+24.1}\\
    \bottomrule
  \end{tabular*}
\end{table*}

\begin{table*}[t]
  \centering
  \caption{Task success rates (\%) on LIBERO with one demonstration per task   (higher is better). “w/ FoRe” applies the FoRe under the corresponding demonstration regime; $\Delta$ is the absolute gain over the SFT baseline in percentage points. }
  \label{tab:libero_one_trajectory}
  \setlength{\tabcolsep}{8pt}
  \renewcommand{\arraystretch}{0.9}
  \begin{tabular*}{\textwidth}{@{\extracolsep{\fill}}lccccc}
    \toprule
    \multirow{2}{*}{\textbf{Model}} &
    \multicolumn{5}{c}{\textbf{LIBERO}} \\
    \cmidrule(lr){2-6}
    &
    \shortstack{\textbf{Spatial}} &
    \shortstack{\textbf{Object}} &
    \shortstack{\textbf{Goal}} &
    \shortstack{\textbf{Long}} &
    \textbf{Avg} \\
    \midrule
    \multicolumn{6}{c}{One-Trajectory SFT}\\
    \midrule
    VLA-Adapter           & 63.6 & 72.6 & 45.4 & 15.8 & 49.4\\
    \quad\textbf{w/ FoRe} & \textbf{84.2} & \textbf{90.6} & \textbf{61.2} & 32.2 & \textbf{67.1} \\
    \quad$\Delta$  & \textcolor{red!70!black}{+20.6} & \textcolor{red!70!black}{+18.0} & \textcolor{red!70!black}{+15.8} & \textcolor{red!70!black}{+16.4} & \textcolor{red!70!black}{+17.7}\\
    \addlinespace[3pt]
    $\pi_{0.5}$           & 59.4 & 62.0 & 36.8 & 27.8 & 46.5\\
    \quad\textbf{w/ FoRe} & 76.8 & 76.2 & 57.0 & \textbf{48.8} & 64.7\\
    \quad$\Delta$  & \textcolor{red!70!black}{+17.4} & \textcolor{red!70!black}{+14.2} & \textcolor{red!70!black}{+20.2} & \textcolor{red!70!black}{+21.0} & \textcolor{red!70!black}{+18.2}\\
    \bottomrule
  \end{tabular*}
\end{table*}

\subsection{Performance on Real-World Tasks}
\begin{figure}[!htbp]
  \centering
  \includegraphics[width=\linewidth]{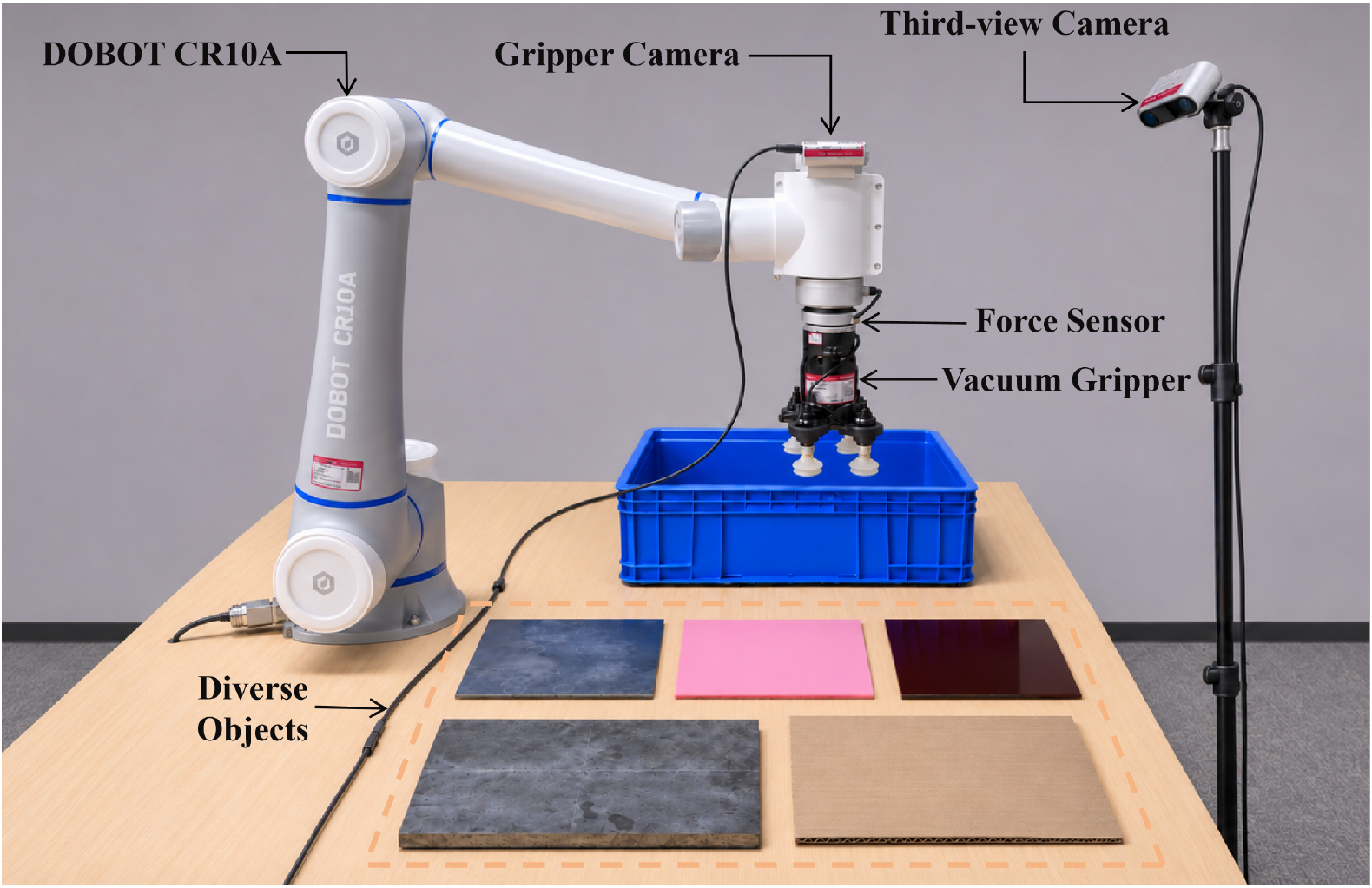}
  \caption{Real-world experimental setup.}
  \label{fig:real environment}
\end{figure}

\paragraph{Experimental setup} We evaluate FoRe on a six-degree-of-freedom DOBOT CR10AF robot with a Robotiq EPick vacuum gripper and Intel RealSense D435i cameras capturing third-person and wrist views (Fig.~\ref{fig:real environment}).

\paragraph{Task categories} We evaluate three task categories with increasing complexity:
(i) \emph{Simple}: pick-and-place tasks without distractors;
(ii) \emph{Cluttered}: pick-and-place tasks with distractors,
requiring target selection based on object identities and
spatial relations; and
(iii) \emph{Long}: multi-object manipulation tasks requiring
sequential execution, such as stacking steel plates, acrylic
sheets, and cardboard panels.

\paragraph{Baselines and evaluation} All policies are trained on real-robot data, with retry policy $M_3$ refined via on-robot RL. We compare $\pi_{0.5}$, VLA-Adapter, and their FoRe-enhanced counterparts over 40 trials per task, reporting success rates under randomized initial object positions.

\begin{figure}[!htbp]
  \centering
  \includegraphics[width=\linewidth]{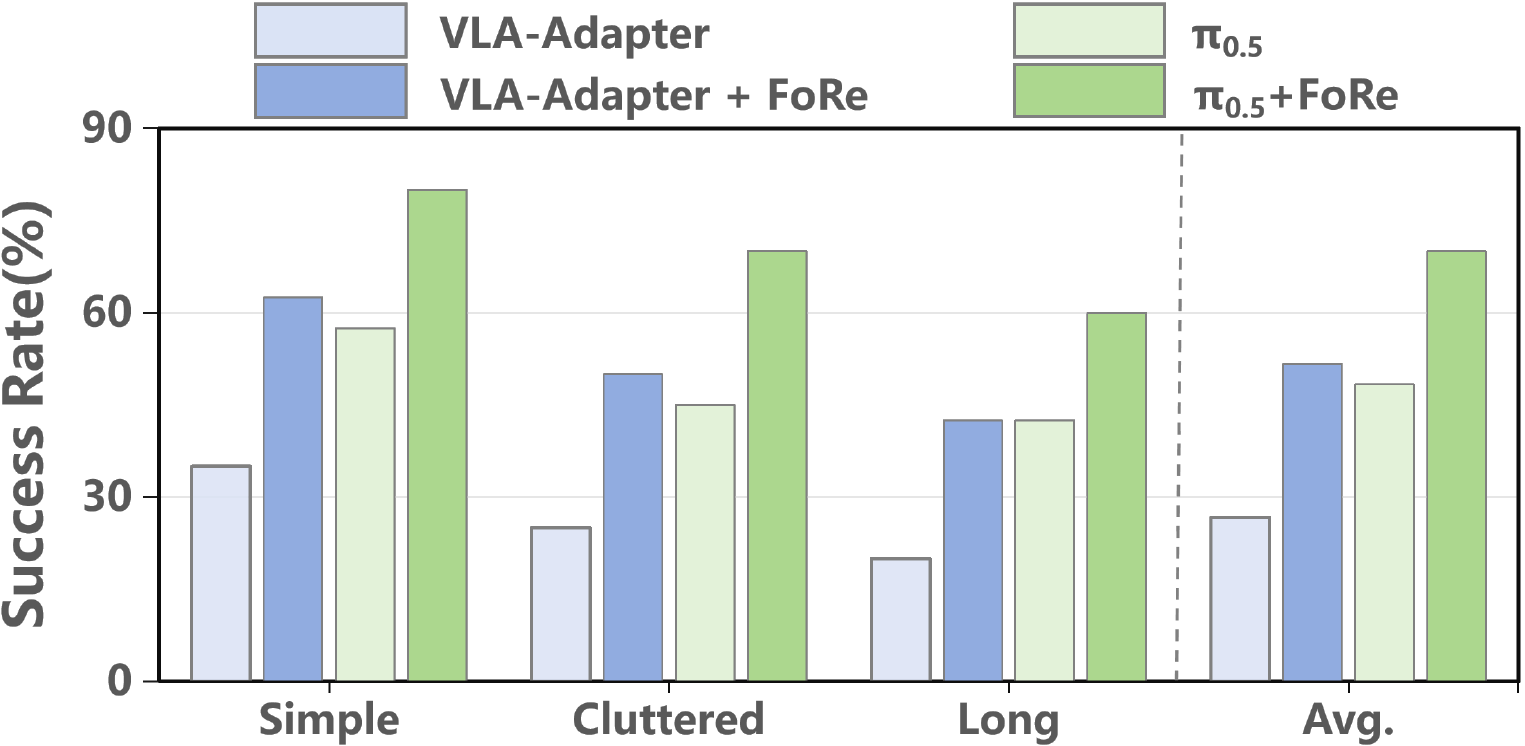}
  \caption{Real-world experimental results.}
  \label{fig:real_world_results}
\end{figure}

\begin{figure}[!htbp]
  \centering
  \includegraphics[width=\linewidth]{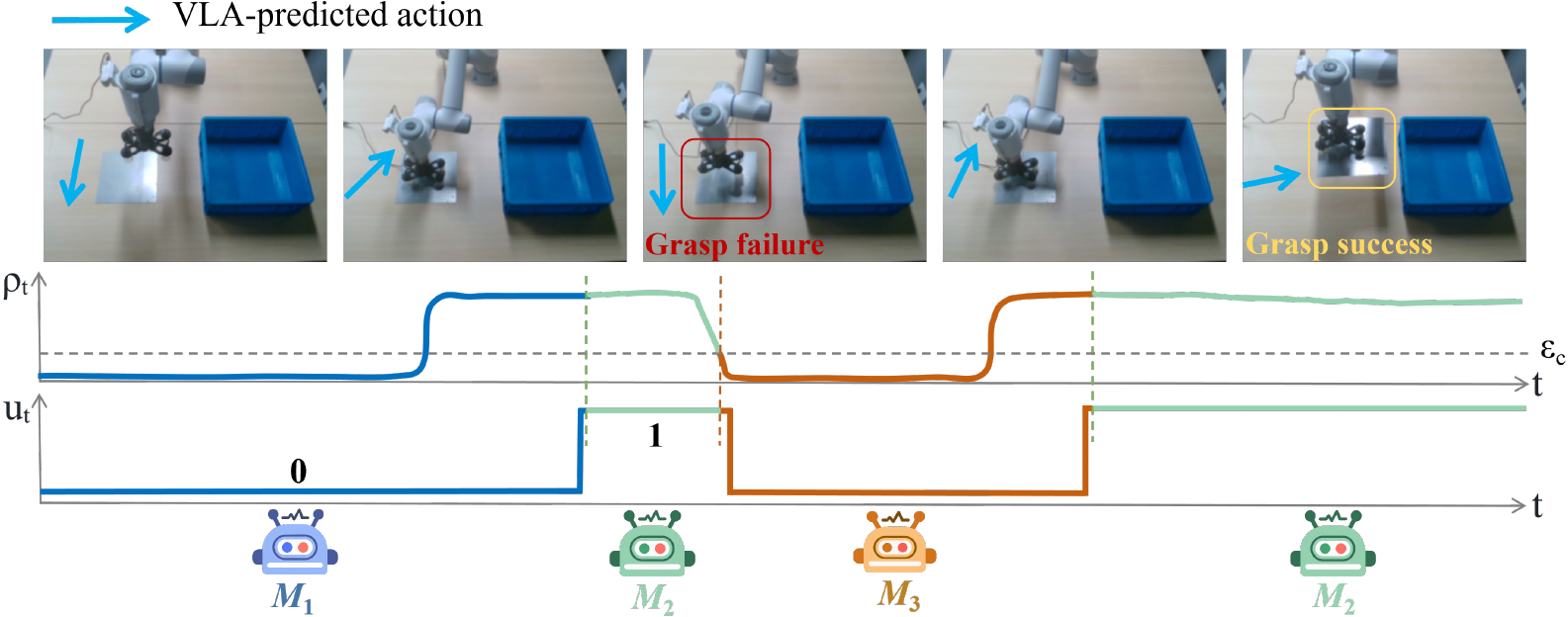}
  \caption{Inference-time grasp recovery on Real-World Tasks. All details are as in Fig.~\ref{fig:LIBERO-SUCTION-visual}.}
  \label{fig:real_world_visual}
\end{figure}

\paragraph{Results}
FoRe improves both backbones across Simple, Cluttered, and Long tasks (Fig.~\ref{fig:real_world_results}). Average success rises from 26.7\% to 51.7\% for VLA-Adapter and from 48.3\% to 70.0\% for $\pi_{0.5}$. On Long tasks, FoRe more than doubles VLA-Adapter's success rate from 20.0\% to 42.5\% and raises $\pi_{0.5}$'s success from 42.5\% to 60\%. Fig.~\ref{fig:real_world_visual} illustrates FoRe's inference-time process for autonomous recovery from local grasp failures in real-world tasks.

\subsection{Performance with Fewer Demonstrations}

\begin{table*}[!t]
  \centering
  \caption{Ablation settings of FoRe with VLA-Adapter on LIBERO-SUCTION.
    \emph{w/o Grasp Retries} disables failure-triggered retries;
    \emph{w/o Retry Policy Optimization} reuses $M_1$ for retries instead of $M_3$ ;
    \emph{w/o Stage-Specific SFT} uses a full-trajectory policy for nominal execution and retry-policy initialization. With trained policies fixed,
    \emph{w/o Gripper-State Conditions} uses force-only switching;
    \emph{w/o Force Switching} uses command-only switching, leaving $M_3$ uninvoked.}
  \label{tab:fore_ablations}
  \setlength{\tabcolsep}{8pt}
  \renewcommand{\arraystretch}{0.9}
  \begin{tabular*}{\textwidth}{@{\extracolsep{\fill}}lcccc}
    \toprule
    \multirow{2}{*}{\textbf{Variant}} &
    \multicolumn{4}{c}{\textbf{LIBERO-SUCTION}} \\
    \cmidrule(lr){2-5}
    &
    \shortstack{\textbf{Suction-Spatial}} &
    \shortstack{\textbf{Suction-Object}} &
    \shortstack{\textbf{Suction-Goal}} &
    \textbf{Avg} \\
    \midrule
    VLA-Adapter + FoRe                  & \textbf{91.3} & \textbf{78.8} & \textbf{95.2} & \textbf{87.6}\\
    \quad{w/o Grasp Retries}            & 81.1 & 64.0 & 86.8 & 76.2 \\
    \quad{w/o Retry Policy Optimization}& 85.8 & 73.0 & 91.2 & 82.4 \\
    \quad{w/o Stage-Specific SFT}       & 84.2 & 65.3 & 90.4 & 78.7 \\
    \quad{w/o Gripper-State Conditions} & 86.7 & 71.0 & 82.8 & 80.1 \\
    \quad{w/o Force Switching}          & 80.2 & 60.8 & 85.2 & 74.3 \\
    \bottomrule
  \end{tabular*}
\end{table*}

\paragraph{Experimental Settings}
On both LIBERO-SUCTION and LIBERO, we compare VLA-Adapter and $\pi_{0.5}$ under One-Trajectory SFT with their FoRe-enhanced counterparts using one demonstration per task. FoRe's stage-specific SFT and RL initial-state sampling use only that trajectory.

\paragraph{Results}
Tables~3 and~4 show that FoRe consistently improves both backbones
with one demonstration per task. On LIBERO-SUCTION, VLA-Adapter's
overall success rate increases from 32.8\% to 55.5\%, while
$\pi_{0.5}$ improves from 39.2\% to 63.3\%, yielding absolute gains
of 22.7\% and 24.1\%, respectively. On LIBERO, the corresponding
success rates increase from 49.4\% to 67.1\% and from 46.5\%
to 64.7\%, with absolute gains of 17.7\% and 18.2\%.
Both backbones improve on every task suite across the two
benchmarks, supporting FoRe's effectiveness under scarce
task-specific demonstrations.

\section{Ablation Experiments}

We ablate FoRe's components and switching rules on LIBERO-SUCTION using VLA-Adapter, the full demonstration set, and the protocol in Sec.~4.1 (Table~\ref{tab:fore_ablations}).

Disabling failure-triggered retries (\emph{w/o Grasp Retries}) while
retaining $M_1$, $M_2$, and their force-guided switching rules reduces
overall success from 87.6\% to 76.2\%. Reusing $M_1$ directly for retries
without RL refinement (\emph{w/o Retry Policy Optimization}) achieves
82.4\%. Thus, the approach--grasp policy alone provides an absolute gain of
6.2\% over disabling retries, indicating a degree of retry capability, while RL refinement contributes a further 5.2\% by adapting this policy to retry scenarios.

Replacing $M_1$ and $M_2$ with a unified policy $P$ trained on complete
trajectories (\emph{w/o Stage-Specific SFT}) reduces overall success to
78.7\%, an absolute decrease of 8.9\%. This variant retains phase
tracking for switching; $M_3$ is initialized from $P$ and optimized
with the same RL procedure and retry-state sampling protocol.
This result supports stage-specific training, which isolates the
approach--grasp policy as the initialization for $M_3$, providing
a better starting point for RL refinement and more effective
grasp retries.

We further examine switching rules while keeping demonstration
annotation and policy training unchanged. Removing the ON/OFF grasp-command conditions while retaining force thresholds
(\emph{w/o Gripper-State Conditions}) reduces overall success to
80.1\%, an absolute decrease of 7.5\%. Here, low force during $M_2$
always invokes $M_3$, making intentional release indistinguishable
from grasp failure and potentially triggering unnecessary retries
that interrupt task progression.

Removing force from  switching rules
(\emph{w/o Force Switching}) reduces overall success to 74.3\%,
an absolute decrease of 13.3\%. ON and OFF commands now trigger
$M_1\!\rightarrow\!M_2$ and $M_2\!\rightarrow\!M_1$, respectively, while $M_3$ is never invoked without force-based failure detection. This result underscores the necessity of force feedback for post-failure recovery.

\section{Conclusion and Limitation}
We presented FoRe, a framework that combines force-guided stage control with local retry learning for autonomous grasp recovery in VLA-based manipulation. We observed continued transport after failed grasps, potentially reflecting overreliance on proprioceptive cues for stage progression and limited retry supervision. To address these issues, FoRe combines separate training of two stage-specific policies with local RL refinement of a dedicated retry policy, using force feedback alongside grasp-command status to detect grasp failures and control stage transitions. Experiments on LIBERO-SUCTION, LIBERO, and real-world tasks show consistent improvements, with average real-world success increasing from 26.7\% to 51.7\% for VLA-Adapter and from 48.3\% to 70.0\% for $\pi_{0.5}$. These findings suggest that force feedback grounds stage control and retry learning in physical grasp outcomes, enabling more reliable autonomous recovery in VLA policies.

However, as an external recovery framework, FoRe relies on the approach–grasp  policy to establish contact with the target object. If the policy consistently fails to make contact with the target, this may impair the ability of FoRe's force-guided controller to advance stages, detect grasp failures, and invoke the retry policy, thereby limiting autonomous grasp recovery. Future work will focus on guiding VLA policies to reliably bring the gripper into contact with the target object.
{
    \small
    \bibliographystyle{ieeenat_fullname}
    \bibliography{main}
}


\end{document}